# Depression Recognition using Remote Photoplethysmography from Facial Videos

Constantino Álvarez Casado, Manuel Lage Cañellas, and Miguel Bordallo López,

***Abstract*— Depression is a mental illness that may be harmful to an individual's health. The detection of mental health disorders in the early stages and a precise diagnosis are critical to avoid social, physiological, or psychological side effects. This work analyzes physiological signals to observe if different depressive states have a noticeable impact on the blood volume pulse (BVP) and the heart rate variability (HRV) response. Although typically, HRV features are calculated from biosignals obtained with contact-based sensors such as wearables, we propose instead a novel scheme that directly extracts them from facial videos, just based on visual information, removing the need for any contact-based device. Our solution is based on a pipeline that is able to extract complete remote photoplethysmography signals (rPPG) in a fully unsupervised manner. We use these rPPG signals to calculate over 60 statistical, geometrical, and physiological features that are further used to train several machine learning regressors to recognize different levels of depression. Experiments on two benchmark datasets indicate that this approach offers comparable results to other audiovisual modalities based on voice or facial expression, potentially complementing them. In addition, the results achieved for the proposed method show promising and solid performance that outperforms hand-engineered methods and is comparable to deep learning-based approaches.**

## I. Introduction

Major depressive disorder (MDD), also known as clinical depression, is a common mental disorder that contributes significantly to the global healthcare burden and can lead to severe consequences for individuals both personally and socially. In addition, several studies suggest long-term and clinically significant depression as a trigger for other serious medical conditions and physiological changes such as cardiovascular disease, diabetes, osteoporosis, aging, pathological cognitive changes, including Alzheimer's disease and other dementias, and even an increase in the risk of earlier mortality [1].

Submitted on: Reviewed on: Published on:

This research has been supported by the Academy of Finland 6G Flagship program under Grant 346208 and PROFI5 HiDyn programunder Grant 326291

All authors are with the Center for Machine Vision and Signal Analysis (CMVS), University of Oulu, 90570 Oulu, Finland. Miguel Bordallo López is also with VTT Technical Research Centre of Finland Ltd, 90571 Oulu, Finland.(e-mail: constantino.alvarezcasado@oulu.fi; manuel.lage@oulu.fi; miguel.bordallo@oulu.fi)

Currently, depression screening is usually based on medical interviews described in the Diagnostic and Statistical Manual of Mental Disorders (DSM-V), but depends on the subjectivity and experience of the psychiatrist and the subjective memory of the patient, a fact that can lead to misdiagnosis with its consequential social, physiological, or psychological side effects due to undertreatment or overtreatment of the illness.

In recent years, the assessment of depression from facial videos has aroused interest in the scientific community, since the clinical literature has documented particular visual cues and behaviors on faces and facial expressions triggered by major depressive disorder [2]. These facial signs go from reducing facial movements, eyebrow activity, eyes gaze, head pose, mood expressions occurrence, body gestures, or eyelid activity, among others. In addition, this discipline allows the development of a noninvasive and unobtrusive technology and modality that can support the medical diagnosis while the physician focuses exclusively on the patient. The literature studies based on facial visual information have concentrated mainly on three ideas: extracting features from textures and dynamic textures using handcrafted textural descriptors, extracting features from the facial geometry and morphology, and using deep learning approaches, which represent the state-of-the-art methods nowadays.

On the other hand, other objective biomarkers have been shown to be useful for physicians to evaluate and assess the level of depression of the patient in a more confident and precise manner. Recent studies have demonstrated the impact of depression on physiological biomarkers, such as heart rate variability (HRV) calculated from the electrocardiogram (ECG) [3] [4], HRV using PPG signals [5] or electrodermal activity (EDA) [6].

In this article, and based on these findings, we propose a novel approach for automatic depression screening using physiological signals extracted from facial videos and machine learning for the first time. Our main contribution can be summarized as follows:

- We assess depression scores by extracting remote photoplethysmographic signals (rPPG), and use them to compute a set of statistical and heart rate variability (HRV) features, including and non-linear geometrical parameters from the blood volume pulse (BVP), feeding them to machine learning regressors based on Random Forests and Multilayer Perceptrons.
- To demonstrate the validity of our approach, we evaluate our methods in two publicly available video-based

datasets, typically used as a benchmark for depression assessment, AVEC2013 and AVEC2014. The results show that the new approach is feasible and shows more stable inter-video predictions than other modalities.

- To complement our study, we compare our approach with different audiovisual modalities. We prove that the combination of physiological signals with both texture-based and deep features is complementary and improves the results further.

## II. Proposed Methodology

In this article, we propose a regression task to determine the level of depression of a person using remote photoplethysmography (rPPG), a technique to extract physiological signals remotely without attaching sensors. In this case, we use rPPG signals extracted from faces recorded with a user-graded RGB camera. The regression task comprises several steps: extracting the biosignals from the facial videos, pre-processing the extracted signals to convert them into physiological rPPG signals, extracting features from the rPPG signals, training the models using those features, and evaluating the performance of the models.

### A. Remote Photoplethysmographic signal extraction

To extract rPPG signals from the videos, we utilize our unsupervised pipeline for the extraction of blood volume pulse signals from faces, called Face2PPG [7].

The unsupervised methodology for remote photoplethysmographic (rPPG) imaging, uses an RGB camera, and it is comprised of several steps: face detection and face alignment, skin segmentation, regions of interest (ROIs) selection, extraction of the raw signals from ROIs, filtering of the raw signals, RGB to PPG transformation and spectral analysis and post-processing to compute vital signs such as heart rate (HR), respiratory rate (RR), blood oxygen saturation (SpO2) or heart rate variability (HRV) [8]. An schematic of the pipeline can be seen in Figure 1.

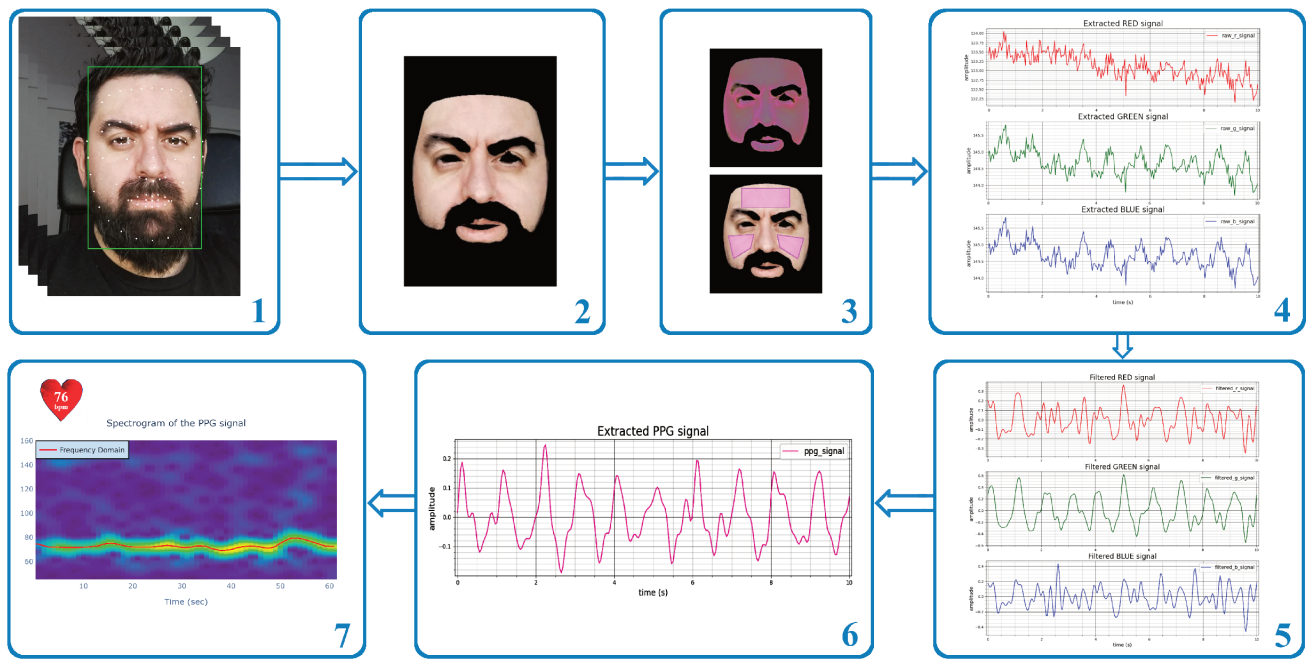

Fig. 1. Unsupervised methodology for remote photoplethysmographic (PPG) imaging using a RGB camera, comprising several steps: 1) Detection and alignment of the face at every frame. 2) Skin segmentation. 3) ROI selection. 4) Extraction of the raw signals from RGB channels at the regions of interest. 5) Filtering of the raw signals on the frequency band of interest. 6) Transformation of the filtered RGB signals to a pulse-type signal. 7) Computation of heart-related features using spectral analysis and post-processing. Figure extracted from our previous work [7]

In particular, our Face2PPG configuration includes the following modules: First, it includes an accurate and robust deep learning-based face detection method based on a Single Shot Multibox Detection network (SSD) [9]. After that, the detected faces are aligned using a deep learning facial landmarks detector named Deep Alignment Network, [10] which gives exceptional performance in terms of accuracy even in challenging conditions [11]. Finally, these landmarks are used in a geometrical skin segmentation and normalization scheme that employs the 85 facial landmark points detected in the face by creating a fixed facial mesh composed of 131 triangles, fixing their coordinates in a normalized frontal pose. The results of the face normalization to extract the biosignals can be seen in Figure 2. The normalized face is processed further using a dynamic multi-region selection scheme that extracts raw RGB signals from the best facial areas, based on statistics. The raw signals are then processed using an improved filtering module that includes detrending and bandpass filtering to remove artifacts and clean the raw signal to the frequency band of interest, and a module to transform the RGB signals into rPPG signals, that includes several reference RGB to PPG conversion methods, from where we choose the most typical chrominance-based method (CHROM) [12]. A detailed explanation of each module can be seen in our previous work [7].

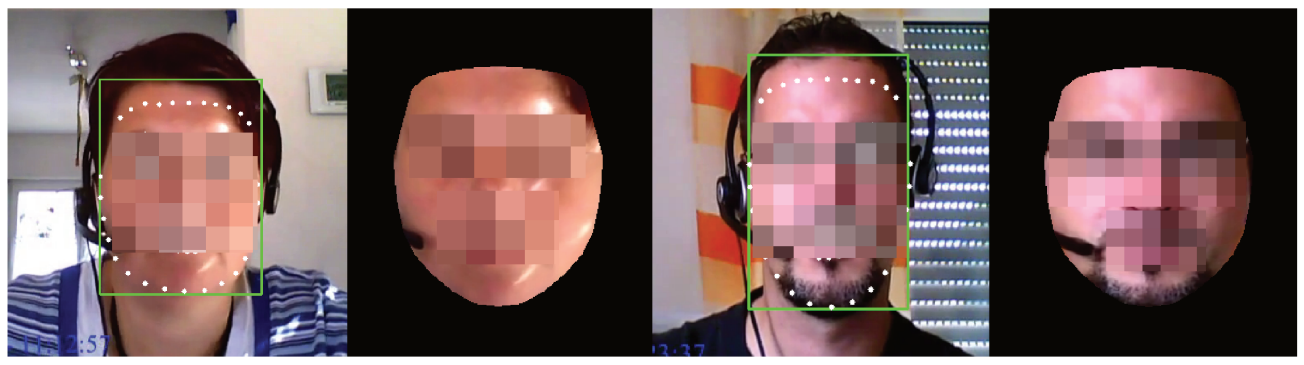

Fig. 2. Normalization of the faces to fixed coordinates of two sample videos from the AVEC2014 database. The left image of each pair shows face detection and landmarks. The right image shows the normalization of each detected face in the videos. [7]. Images pixelated for privacy reasons.

### B. Feature extraction

To train our regression models, we use rPPG signals extracted from visual information to compute 68 features along different windows of each 1-dimensional signal. We used windows of 6 seconds and a fixed sliding window of 0.33 seconds, which is equivalent to 10 video frames for a typical framerate of 30 fps. An example rPPG signal window, is shown in Figure 3. For each rPPG signal window, the extracted features include nine statistical features for time-series, six fractal analysis features, six entropy analysis features, and 49 heart-related features in time-domain, frequency-domain, and non-linear features, extending the 30 features used in our previous related work [13].

In particular, the statistical features, include the *mean, min, max, std, dynamic range* and four *percentiles* (10, 25, 75 and 90) [14]. The fractal analysis features include the *Katz fractal dimension, Higuchi fractal dimension* and *detrended fluctuation analysis* of the entire window, and the mean of the three fractal analysis features computed in sub-windows of 2 seconds of the whole window. The entropy analysis features include *permutation entropy, spectral entropy, approximate entropy, sample entropy, Hjorth mobility and complexity* and *number of zero-crossings* of the entire window. The heart and HRV related features include heart rate (HR), breathing rate (BR), interbeat interval (IBI), differences between R-R intervals (pNN20, pNN50), Poincare analysis, frequency domain components (VLF, LF, HF, LF/HF ratio), the standard deviation of NN intervals (SDNN), among others [15]. To compute them, we use the *Numpy Python* library to compute the statistical features, the *Antropy Python* package, a software tool for computing the complexity of time-series, to extract both fractal and entropy features [16] and two Python libraries, namely *Neurokit2* [17] and *HeartPy* [18], to computer HRV related features.

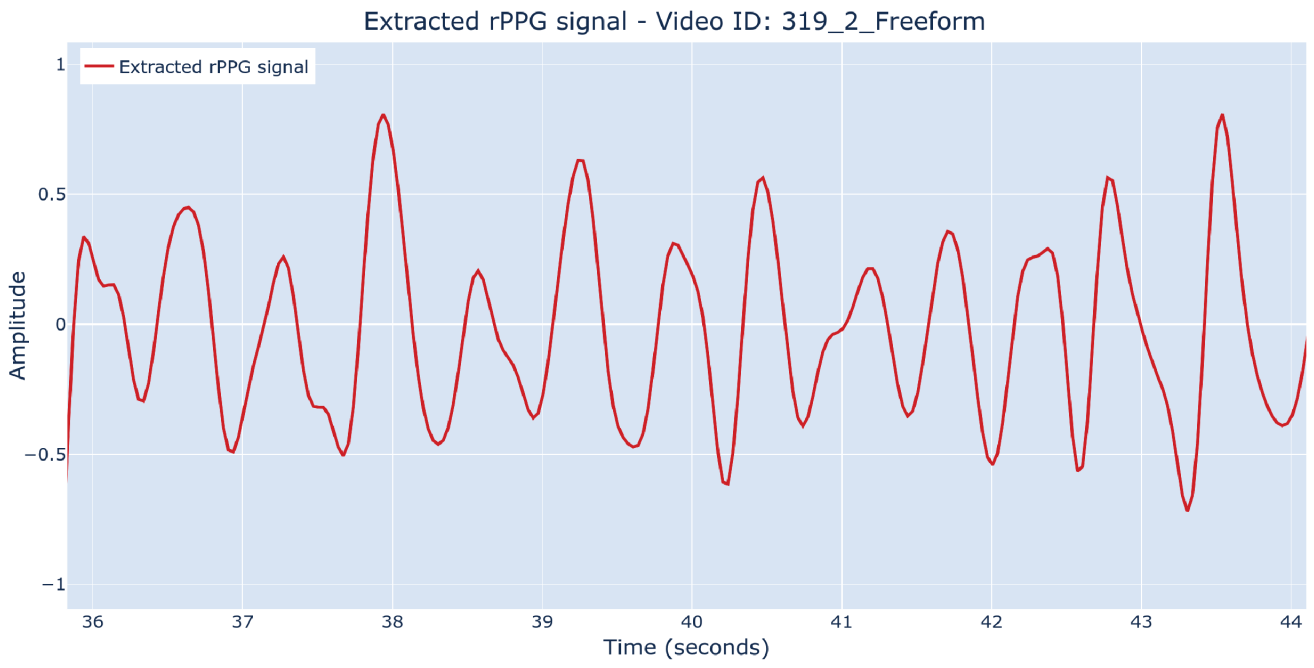

Fig. 3. An example rPPG signal window extracted from a video included in the AVEC2014 dataset.

For comparative purposes, we have also computed textural features from visual information. We have followed a similar approach to the AVEC2014 baseline [19], which employs the local dynamic appearance descriptor LGBP-TOP [20], employing fixed temporal windows of 10 consecutive frames. Following the baseline method, the extracted feature vector comprises features extracted only from the XY orthogonal plane. The computation of textural features employs a custom Python script based on the Bob signal processing and machine learning library [21].

## C. Regressor selection

For both physiological and textural features, we select regressors based on Random Forests and Multilayer Perceptrons, as included in *Scikit-learn* Python library. The Random Forest Regressor (RFR) uses $n_estimator = 550$, $max_depth = 15$ and default values for the rest of the parameters of the model. The Multilayer Perceptron Regressor (MLPR) uses a topology that includes an input layer with the number of input features, three hidden layers, and an output layer with one neuron that corresponds to the regression value of the depression. The configuration used for the training includes: a *"relu"* (rectified linear unit function) for the activation function in the hidden layers, *"Adam"* solver for the weight optimization, a $batch_size = 140$ with a learning rate *"constant"*, an initial learning rate of 0.01 and default values for the rest of the parameters.

Again, for comparative purposes, we have implemented an end-to-end deep-learning regression model based on a ResNet-50 convolutional neural network, followed by a regression layer composed of two fully connected layers. Based on the literature [22], as input to the network, we have used all individual frames of each video by cropping the input frame to the facial rectangle.

We evaluate the performance of the regression models both individually and combined. First, we train individual models using extracted features from the rPPG physiological signals, and compare them with the performance of regressors based on textural features and the end-to-end regressor based on deep-learning. In addition, we combine these features and models in two different ways. First, using a feature-level fusion approach (pre-fusion) by creating a unique feature vector with features from both textural and physiological modalities, training a model with these feature vectors. Finally, we also use a score-level fusion approach (post-fusion) by combining the result of the inferences from the individual models using the average of the results.

# III. EXPERIMENTAL ANALYSIS

## A. Datasets and protocol

To demonstrate the performance of the proposed method, we evaluate the trained models on two publicly available databases, namely the Audio/Visual Emotion Challenge (AVEC) 2013 [23] and 2014 [24]. The experiments were performed on the sets of the Depression Recognition Sub-Challenge (DSC) task, where the goal was to estimate the score of individuals on the Beck Depression Inventory (BDI-II). Both datasets are derived from a subset of the audiovisual depressive language corpus (AViD-Corpus) and they are divided in three partitions: training, development, and test set. Every video includes a label based on questionnaire answers following the Beck Depression Inventory-II (BDI-II) [2], resulting in a depression score of 0 to 63. According to the BDI-II score, the severity of depression can be classified into four levels: minimal (0-13), mild (14-19), moderate (20-28), and severe (29-63).

The AVEC2013 dataset contains 150 videos from 84 subjects, with 50 videos on each partition. However, in the AVEC2014 dataset, the individuals were recorded while performing two different tasks: Freeform and Northwind. The recordings are segmented into three parts in both tasks: training, development, and test set containing 50 videos in each partition for a total of 300 videos. The protocol for AVEC2014 evaluates the models using the two different tasks, both separately and jointly. For the separate task models, models are trained using the subsets of either the *Northwind* or *Freeform* tasks, while the joint models, simply combine the data from both tasks both in the training and testing phases.

## B. Experimental setup

We evaluated and analyzed the proposed methodologies to detect the level of depression using features extracted from remote photoplethysmography signals and visual features extracted from video frames from both benchmark data sets. We compare the results across different trained models using these features individually or in a fusion manner and compare them with state-of-the-art for both supervised and unsupervised methods. The experiments are performed using a computer that includes an AMD® Ryzen(TM) 3700X 8-core processor at 3.6GHz, with 64 Gigabytes of RAM, 4 terabyte SSD and two NVIDIA GeForce® RTX(TM) 2080. We have also used the *Puhti* supercomputer at the IT Center for Science (CSC) in Finland to extract the visual texture-based features. We used Python 3.8 as the programming language.

## C. Performance metrics

To evaluate the performance of these models and make a fair comparison with the state-of-the-art methods, we provide the two most common metrics in the automatic depression assessment literature, Mean Absolute Error (MAE) and Root Mean Square Error (RMSE). The overall predicted depression score for each input video is obtained by averaging the estimation scores for all its windows.

## D. Experimental results

In this section, we evaluate the performance and validity of the proposed modality and approach through a series of experiments in the benchmark databases. We compare them with other modalities and state-of-the-art approaches.

*1) Performance in AVEC2013 and AVEC2014:* In Table I and Table II, we show the evaluation of the performance of the proposed approach using HRV and BVP features extracted from facial videos for both AVEC2013 and AVEC2014. In addition, we also explore a multimodal fusion by combining the heart-related features with textural and deep features to complement the results.

We observe that the results of the individual models (using HRV features and textural features) on the AVEC2013 and AVEC2014 test sets have similar performance, although slightly improved when using features from temporal visual descriptors based on textures. The most remarkable output is the combination of the features from both textural and physiological modalities, achieving the best results among those not based on deep learning features.

TABLE I

PERFORMANCE OF THE PROPOSED METHOD AND MODELS FOR DEPRESSION RECOGNITION ON AVEC2013, MEASURED IN MEAN ABSOLUTE ERROR (MAE) AND ROOT MEAN SQUARE ERROR (RMSE) OVER ALL SEQUENCES. WE USE THE FOLLOWING NOTATION TO REFER TO THE MACHINE LEARNING ALGORITHMS: RFR FOR RANDOM FOREST REGRESSOR AND MLPR FOR A MULTILAYER PERCEPTRON REGRESSOR.

| Features | Fusion type | Model | MAE | RMSE |
|---|---|---|---|---|
| rPPG | - | RFR | 7.97 | 9.98 |
| rPPG | - | MLPR | 7.54 | 9.75 |
| Textural | - | MLPR | 7.26 | 8.99 |
| rPPG + Textural | Pre | MLPR | 6.98 | 9.02 |
| rPPG + Textural | Post | RFR + MLPR | 7.03 | 8.97 |
| rPPG + Textural | Post | MLPR + MLPR | **6.43** | **8.01** |

For AVEC2014, we can observe that for the *Freeform* task the regression models work slightly better than for the *Northwind* task, as expected according to the baseline results

TABLE II

PERFORMANCES OF THE PROPOSED METHODS FOR DEPRESSION RECOGNITION CONSIDERING SINGLE TASK AND FUSION OF TASKS ON AVEC2014. PERFORMANCE IS MEASURED IN MEAN ABSOLUTE ERROR (MAE) AND ROOT MEAN SQUARE ERROR (RMSE) OVER ALL SEQUENCES.

| Features | Fusion | Model | Task | MAE | RMSE |
|---|---|---|---|---|---|
| rPPG | - | RFR | Freeform | 7.74 | 9.68 |
| Textural | - | MLPR | Freeform | 7.43 | 9.33 |
| rPPG + Textural | Pre | RFR | Freeform | 8.03 | 9.84 |
| rPPG + Textural | Post | RFR + MLPR | Freeform | 7.37 | 8.72 |
| rPPG | - | RFR | Northwind | 8.28 | 10.76 |
| Textural | - | MLPR | Northwind | 8.17 | 10.40 |
| rPPG + Textural | Pre | RFR | Northwind | 7.21 | 8.99 |
| rPPG + Textural | Post | RFR + MLPR | Northwind | 7.62 | 9.64 |
| rPPG | - | RFR | Fusion | 7.44 | 9.55 |
| Textural | - | MLPR | Fusion | 7.02 | 9.08 |
| Deep | - | ResNet-50 | Fusion | 6.83 | 9.06 |
| rPPG + Textural | Pre | RFR | Fusion | 7.20 | 9.03 |
| rPPG + Textural | Post | RFR + MLPR | Fusion | 6.81 | 8.63 |
| rPPG + Deep | Post | RFR + ResNet-50 | Fusion | 6.90 | 8.88 |
| All | Post | All | Fusion | **6.57** | **8.49** |

[24]. We can observe that the results of the individual models (using HRV features and textural features individually) when using the data joining both tasks are similar in both datasets.

We show both results for individual modalities and the fusion of HRV features with both textural and deep features.In AVEC14, score-level fusion also results in better performance than feature-level fusion although slightly worst than in AVEC2013.

Although we can observe that the deep learning-based approach (ResNet-50) has better individual results than the models trained with handcrafted features extracted from either textural or rPPG features, its combination with rPPG features at score-level shows a further improvement of the results. This proves that, in the same manner as textural and rPPG modalities, deep models provide for information that is also complementary to that extracted from physiological signals. However, the best results are obtained when fusing all three data modalities at score level.

*2) Error analysis:* To further analyze the performance of the features we show the error distribution in AVEC2014 benchmark, and show it in Figure 4. The figure shows the absolute error for each of the 100 test videos ordered from the smallest to the largest. We can observe that the error distributions of both the HRV and the deep models have similar shapes. While the deep model shows an overall smaller error, the HRV shows a smaller maximum error, always smaller than 25. In contrast, the textural model shows a higher portion of videos with a very small error (error 10 or less), but jumps to very high errors for a significant portion of the videos.

*3) Qualitative evaluation:* For a qualitative evaluation of the models, we show the different predictions per window for three different example videos, depicted in Figure 5. We can observe that inference when using rPPG-based features to train the models is relatively stable and shows less variance for the different time windows that make up a single video. This is in contrast with the the inferences obtained from regressors trained with visual textural features, that show high variability in the predictions, although a somehow accurate average. Models trained using deep learning, show a reasonable stability, but worse than HRV.

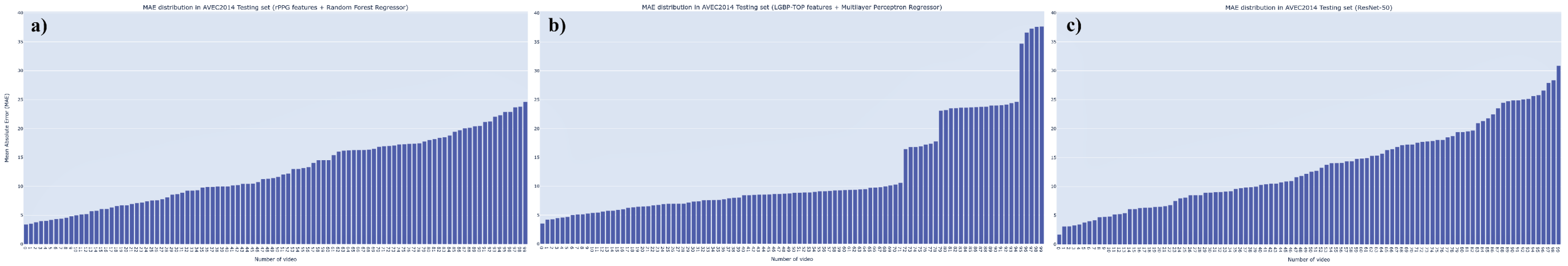

Fig. 4. Mean Absolute Error (MAE) distribution of the AVEC2014 Testing Video dataset (Northwind + Freeform). Error distribution of the depression ordered from smallest to largest error per video. From left to right: a) error distribution when using: rPPG+HRV features + Random Forest regressor, b) LGBP-TOP features + Multilayer Percepton regressor and c) ResNet-50 neural network.

*4) Impact of the Window size:* We compare the results of the proposed method using different window sizes to extract HRV features from the rPPG signals and a fixed sliding window of 0.33 seconds (10 video frames). We have carried out this experiment in AVEC2014 using the same Random Forest regressor as in Table II and the data from both tasks included in AVEC2014 (*Freeform* and *Northwind* data). We have tested on typical values five different window lengths: 5, 6, 8, 10, and 15 seconds. The summary of the results can be seen in Table III. The results show that shorter windows that capture short term temporal changes shows a better performance than longer ones, while windows below 6 seconds, start showing problems worse performance due to the lack of sufficient pulse peaks to compute reliable statistics, especially when the subjects have a low heart rate.

TABLE III
PERFORMANCE OF THE REGRESSION MODEL IN AVEC2014 TRAINED WITH HRV FEATURES EXTRACTED FROM THE RPPG SIGNAL USING DIFFERENT WINDOW SIZES (IN SECONDS) AND A FIXED SLIDING WINDOW OF 0.33 SECONDS (10 VIDEO FRAMES).

| | Window Size | | | | |
|---|---|---|---|---|---|
| Metrics | $w = 5$ | $w = 6$ | $w = 8$ | $w = 10$ | $w = 15$ |
| RMSE | 10.54 | 9.55 | 10.24 | 9.94 | 10.29 |
| MAE | 8.36 | 7.44 | 8.13 | 7.92 | 8.47 |

*5) Cross-database analysis:* To observe the generalizability of machine learning models when using physiological features extracted from rPPG signals in the training, Table IV shows the results after performing a cross-database analysis. In this experiment, we train two different machine learning regression models (Random Forest regressor and Multilayer Perceptron regressor) using the training protocol suggested in the source dataset. We tested the resulting models in the *Testing* subset of the target database. In the first row of the table, we depict the results of two models trained in AVEC2013, and tested in the testset of AVEC2014. The second row shows the vicecersa results, using AVEC2014 to training the models, and AVEC2013 to test them.

We can observe that the results are comparable to the ones obtained in Table I and in Table II, showing a good generalization capability when using this type of features as discriminative features to estimate the level of depression from facial videos.

TABLE IV
PERFORMANCE OF THE PROPOSED METHOD USING RPPG FEATURES AND TWO DIFFERENT REGRESSION MODELS IN CROSS-DATASET SETTING.

| | | RF regressor | | MLP regressor | |
|---|---|---|---|---|---|
| Training set | Test set | MAE | RMSE | MAE | RMSE |
| AVEC2013 | AVEC2014 | 7.52 | 9.48 | 7.07 | 9.94 |
| AVEC2014 | AVEC2013 | 7.45 | 9.64 | 7.90 | 9.98 |

*6) Performance across different machine learning regression models:* We explore the performance across different regression models and summarize the results in Table V. We have trained a set of Machine Learning regressors selected using an exploratory strategy that tried up to 15 different regressors, which we narrowed down to 6 based on their type and preliminary performance. We selected Random Forest regression (RFR) and Extremely Randomized Trees regression (ExTR) from ensemble learning methods, Logistic regression (LogR) and Support Vector Machine regression (SVR) as linear regressors, Stochastic Gradient Descent regression (SDGR) as iterative method and Multilayer Perceptron regression (MLPR) as neural network method. For each model, we have used the default parameters of the machine learning algorithms set by the *Scikit-learn* Python library, with the exception of an increased number of estimators and maximum depth for the models based on trees.

Similarly to the experiments shown in Table I and Table II, we explore the results when training the different models with visual and rPPG features individually, and using two multimodal fusion approaches.

We can observe that in general the Random Forest regressor and the Multilayer Perceptron regressor obtain the best results. The RFR works especially well when using the features extracted from the rPPG signals. The MLPR works especially well when using the visual features. We hypothesize that in the case of the HRV features, the RFR is able to find nonlinear relationships between the dependent and independent variables whereas the MLPR works better with linear relationships, assuming that the features extracted from dynamic textures of a face have a strong linear dependency. The logistic regressor

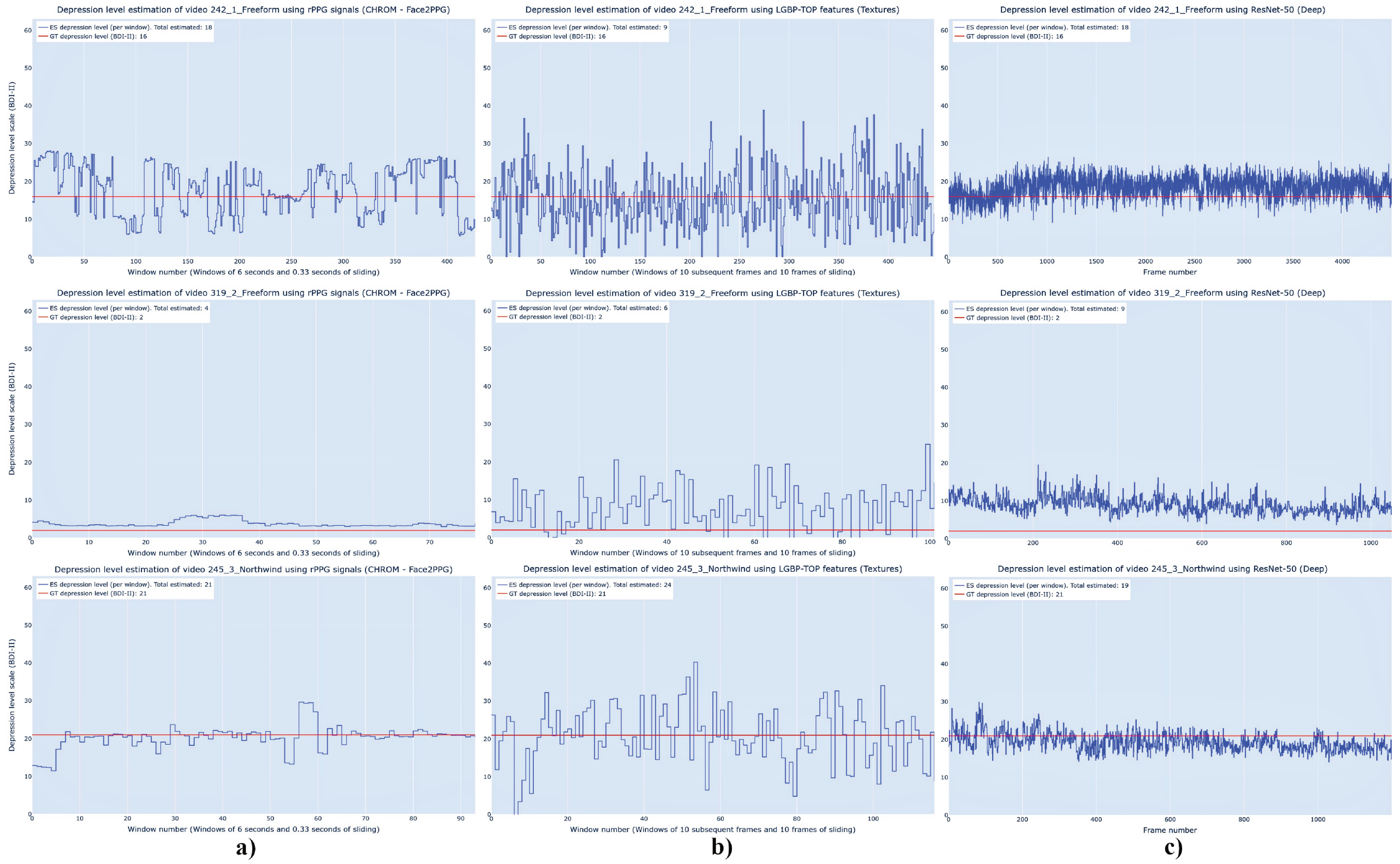

Fig. 5. Examples of the predicted depression level per window in three videos from the AVEC2014 test set. In the first row, estimation for the video *319_2* performing the *Freeform* task, in the second row, the estimation for the video *245_3* performing the *Northwind* task, and in the third row, estimation for a long video (*242_1* performing the *Freeform* task. From left to right, estimations of: a) Random Forest regressor using rPPG features, b) Multilayer Perceptron regressor using the visual textural features, and c) ResNet-50 trained with input facial images.

TABLE V

PERFORMANCE OF DIFFERENT REGRESSION MODELS IN AVEC2014 USING RPPG FEATURES,,VISUAL FEATURES (LGBP-TOP) FEATURES AND THE FUSION OF BOTH AT FEATURE-LEVEL (PRE-FUSION) OR SCORE-LEVEL(POST-FUSION).

| Metric | Features | regression model | | | | | |
|---|---|---|---|---|---|---|---|
| | | RF | ExTR | LogR | SVR | SDGR | MLPR |
| MAE | rPPG | **7.44** | 8.18 | 10.69 | 8.91 | 9.15 | 7.94 |
| | Visual | 8.15 | 7.99 | 7.85 | 7.92 | 8.21 | **7.18** |
| | rPPG + Visual (Post) | 7.66 | 7.89 | 8.42 | 8.17 | 8.41 | **7.09** |
| | rPPG + Visual (Pre) | **7.22** | 7.52 | 8.54 | 7.98 | 8.56 | 7.44 |
| RMSE | rPPG | **9.55** | 9.97 | 14.71 | 11.12 | 11.24 | 10.36 |
| | Visual | 9.96 | 9.61 | 10.71 | 10.08 | 10.37 | **9.39** |
| | rPPG + Visual (Post) | 9.57 | 9.55 | 11.45 | 10.26 | 10.27 | **8.83** |
| | rPPG + Visual (Pre) | **9.02** | 9.27 | 10.71 | 9.89 | 10.61 | 9.11 |

works well when using the LGBP-TOP features but achieves poor performance when using the HRV features. As expected, extra-trees ensemble regressor has similar performance than the Random Forest, but slightly worst when using rPPG features and slightly better with the LGBP-TOP features, especially for the RMSE metric.

### *E. Comparison of features and sensor modalities*

We have compiled a series of previous works for each modality from baseline to state-of-the-art methods. The primary sensor modalities are based on the typically available sensor modalities such as audio and RGB video, as for AVEC2013 and AVEC2014 database benchmarks. However, the main differences are related to the type of information of interest and the way of computing features from it. Since we introduced a data and feature modality extracted from a remote facial video to regress the level of depression, namely remote physiological features from visual information, we focus on these comparisons. Table VI shows a comparison of different approaches, sensors, and data modalities to infer depression levels in an unobtrusive manner automatically from audiovisual material. We have identified five types of features extracted from both audio and video sensors.

From the audio sensor, previous works have employed features extracted from:

- Speech signals as an audio time series. We have identified features such as handcrafted speech features (LLDs, MFCCs, statistical features, spectral features, etc.), deep learning features, or the conversion to spectral images to extract deep learning visual features.
- Speech as semantic information. Features such as linguistic and para-linguistic features or emotion recognition features.

From the RGB videos, we have identified in the literature four different data (feature) modalities:

- Geometrical features, mostly associated with motion and

TABLE VI

COMPARISON OF DIFFERENT SENSOR AND DATA (FEATURE) MODALITIES FOR DEPRESION ESTIMATION FROM AUDIOVISUAL DATA. NOTATION: TCN: TEMPORAL CONVOLUTIONAL NETWORK, SVM: SUPER VECTOR MACHINE, SVR: SUPER-VECTOR REGRESSOR, MLP: MULTILAYER PERCEPTRON, DCNN: DEEP CONVOLUTIONAL NEURAL NETWORK 2DCNN: 2-DIMENSIONAL CONVOLUTIONAL NEURAL NETWORK, STA: SPATIO-TEMPORAL ATTENTION, EEP: EIGEN EVOLUTION POOLING, LR: LINEAR REGRESSION, PLS: PARTIAL LEAST SQUARE REGRESSION.

| Sensor modality | Features type | Feature Extraction | Year | Method approach | Method | MAE | RMSE | Test dataset |
|---|---|---|---|---|---|---|---|---|
| Audio | Speech | Handcrafted | 2013 | Speech features (Baseline) | Valstar et al. [23] | 10.35 | 14.12 | AVEC2013 |
| Audio | Speech | Handcrafted | 2014 | Speech features (Baseline) | Valstar et al. [24] | 10.04 | 12.57 | AVEC2014 |
| Audio | Speech | Handcrafted | 2018 | MFCC + LR | Jan *et al.* [25] | 8.07 | 10.28 | AVEC2014 |
| Audio | Speech | Deep Learning | 2020 | Spectrum images + STA network | Niu et al. [26] | 7.14 | 9.50 | AVEC2013 |
| Audio | Speech | Deep Learning | 2020 | Spectrum images + STA network | Niu et al. [26] | 7.65 | 9.13 | AVEC2014 |
| Audio | Speech | Deep Learning | 2021 | Speech signal + Spectrum images + ResNet | Dong et al. [27] | 7.32 | 8.73 | AVEC2013 |
| Audio | Speech | Deep Learning | 2021 | Speech signal + Spectrum images + ResNet | Dong et al. [27] | 6.80 | 8.82 | AVEC2014 |
| Audio | Speech | Deep Learning | 2021 | Attention TCN-based (TDCA-Net) | Cai et al. [28] | 6.90 | 9.22 | AVEC2013 |
| Audio | Speech | Deep Learning | 2021 | Attention TCN-based (TDCA-Net) | Cai et al. [28] | 7.08 | 8.90 | AVEC2014 |
| RGB Video | Geometrical | Deep Learning | 2018 | Motion + AlexNet (Landmarks Motion History, Motion History Image, Gabor Motion History) | S'adan et al. [29] | n/a | n/a | AVEC2014 |
| RGB Video | Geometrical | Deep Learning | 2020 | Spectral heatmaps and vectors + CNN + ANN | Zhu et al. [30] | 6.16 | 8.10 | AVEC2013 |
| RGB Video | Geometrical | Deep Learning | 2020 | Spectral heatmaps and vectors + CNN + ANN | Zhu et al. [30] | 5.95 | 7.15 | AVEC2014 |
| RGB Video | Geometrical | Handcrafted | 2022 | Facial landmarks motion + SVM (Landmarks Motion Magniture, Gaze, Action Units) | Rathi et al. [31] | n/a | n/a | DAIC-WOZ |
| RGB Video | Texture | Handcrafted | 2013 | LPQ-TOP + $\epsilon$-SVR (Baseline) | Valstar et al. [23] | 10.88 | 13.61 | AVEC2013 |
| RGB Video | Dynamic texture | Handcrafted | 2014 | LGBP-TOP + SVR (Baseline) | Valstar et al. [24] | 8.86 | 10.86 | AVEC2014 |
| RGB Video | Dynamic texture | Handcrafted | 2015 | Facial LBQ-TOP + SVR | Wen et al. [32] | 8.22 | 10.27 | AVEC2013 |
| RGB Video | Textures | Deep Learning | 2017 | Facial Apparence + DCNN | Zhu et al. [33] | 7.88 | 10.19 | AVEC2013 |
| RGB Video | Textures | Deep Learning | 2017 | Facial Apparence + DCNN | Zhu et al. [33] | 7.82 | 10.36 | AVEC2014 |
| RGB Video | Textures | Deep Learning | 2019 | Facial + ResNet-50 | Melo et al. [34] | 6.30 | 8.25 | AVEC2013 |
| RGB Video | Textures | Deep Learning | 2019 | Facial + ResNet-50 | Melo et al. [34] | 6.15 | 8.23 | AVEC2014 |
| RGB Video | Dynamic Textures | Deep Learning | 2020 | Facial + Two-stream 2DCNN | Melo et al. [22] | 5.96 | 7.97 | AVEC2013 |
| RGB Video | Dynamic Textures | Deep Learning | 2020 | Facial + Two-stream 2DCNN | Melo et al. [22] | 6.20 | 7.94 | AVEC2014 |
| RGB Video | Dynamic texture | Deep Learning | 2021 | Facial 3DCNN features + SVR | Niu et al. [35] | 6.19 | 8.02 | AVEC2013 |
| RGB Video | Dynamic texture | Deep Learning | 2021 | Facial 3DCNN features + SVR | Niu et al. [35] | 6.14 | 7.98 | AVEC2014 |
| RGB Video | Dynamic texture | Deep Learning | 2021 | Upper body images + CNN AlexNet | Ahmad et al. [36] | 5.64 | 7.28 | AVEC2013 |
| RGB Video | Physiological | Handcrafted | 2022 | rPPG and HRV features + RF | Ours | 7.54 | 9.75 | AVEC2013 |
| RGB Video | Physiological | Handcrafted | 2022 | rPPG and HRV features + RF | Ours | 7.44 | 9.55 | AVEC2014 |
| Multimodal | Speech + Textures | Handcrafted | 2013 | Speech features + LBP + PLS | Meng et al. [37] | 9.14 | 11.19 | AVEC2013 |
| Multimodal | Speech + Dynamic textures | Handcrafted | 2014 | Speech features + LGBP-TOP + SVR | Valstar et al. [24] | 7.89 | 9.89 | AVEC2014 |
| Multimodal | Geometrical + Textures | Handcrafted | 2014 | Geometrical features + LPQ + k-NN | Kaya et al. [38] | 7.86 | 9.72 | AVEC2013 |
| Multimodal | Speech + Dynamic textures | Deep Learning + Handcrafted | 2018 | MFCC + VGG-Face features + PLS | Jan et al. [25] | 6.14 | 7.43 | AVEC2014 |
| Multimodal | Speech + Dynamic textures | Deep Learning | 2020 | Speech spectrum images + Facial + STA network + EEP | Niu et al. [26] | 6.14 | 8.16 | AVEC2013 |
| Multimodal | Speech + Dynamic textures | Deep Learning | 2020 | Speech spectrum images + Facial + STA network + EEP | Niu et al. [26] | 5.21 | 7.03 | AVEC2014 |
| Multimodal | Phisiological + Dynamic textures | Handcrafted | 2022 | rPPG and HRV features + LGBP-TOP + RF + MLP | Ours | 6.43 | 8.01 | AVEC2013 |
| Multimodal | Phisiological + Dynamic textures | Handcrafted | 2022 | rPPG and HRV features + LGBP-TOP + RF + MLP | Ours | 6.81 | 8.63 | AVEC2014 |

morphology of both the image and the facial landmarks. The approaches and methods that use these features focus primarily on translating the temporal information of the landmarks or head pose to images such as spectral heat maps, motion history images or motion maps. But other approaches use temporal and morphological information and facial landmark features, gaze, or Action Units (AU) to regress the level of depression.

- Texture features, mostly associated with the static visual features of only one frame. The approaches and methods use handcrafted visual descriptors such as LPQ or LBP features or deep learning features based on the facial appearance of one frame to infer an instantaneous level of depression from the appearance.
- Dynamic texture features include the temporal information based on visual features from a sequence of frames. This is the most explored feature modality since it is known that temporal facial reactions or expressions throw more information about a person's emotional state. The approaches focused on this modality have explored different features such as handcrafted spatio-temporal visual descriptors (LGBP-TOP, LBQ-TOP), different deep learning architectures that encode temporal information, or low-level deep learning features extracted from sequences of images.
- And finally, to the best of our knowledge, we have introduced a new data (feature) modality that can be used on RGB videos. It consists on the extraction of physiological signals (BVP) from faces using the temporal RGB information. We use remote photoplethysmographic waveforms to extract features related to the pulse signal, such as heart rate variability and fractal analysis, which have been shown to have a significant impact on the monitoring and diagnosis of mental health disorders such as depression, stress, or anxiety.

From the comparative results, it can be seen that visual information seems to offer better cues for the assessment of depression than audio information. In particular, deep features that combine both spatial and temporal information offer the best overall performance, while other modalities such as geometrical features, behavioural signals and remote physiological signals (HRV) could offer complementary information, further improving the performance. For audio, deep models also outperform those created using handcrafted features. Overall, the multimodal combination of both audio and video shows the best individual performance.

TABLE VII
COMPARISON OF METHODS FOR PREDICTING THE LEVEL OF DEPRESSION ON THE AVEC2013 DATASET.

| Methods | MAE | RMSE |
|---|---|---|
| AVEC2013 Video Baseline [23] | 10.88 | 13.61 |
| MHH + LBP (Meng *et al.* [37]) | 9.14 | 11.19 |
| LPQ + SVR (Kächele *et al.* [39]) | 8.97 | 10.82 |
| LPQ-TOP + MFA (Wen *et al.* [40]) | 8.22 | 10.27 |
| LPQ + Geo (Kaya *et al.* [38]) | 7.86 | 9.72 |
| Two DCNN (Zhu *et al.* [41]) | 7.58 | 9.82 |
| C3D (Jazaery *et al.* [42]) | 7.37 | 9.28 |
| ResNet-50 (Melo *et al.* [34]) | 6.30 | 8.25 |
| Four DCNN (Zhou *et al.* [43]) | 6.20 | 8.28 |
| 3DCNN + SVR (Niu *et al.* [35]) | **6.19** | 8.02 |
| Two-stream 2DCNN (Melo *et al.* [22]) | **5.96** | **7.97** |
| Ours (HRV) | 7.54 | 9.75 |
| Ours (HRV + LGBP-TOP) | 6.43 | 8.01 |

## F. Comparison with previous work

For modalities based only on visual information, we compare the results of our proposed method against state-of-the-art methods on AVEC2013 and AVEC2014 datasets and show them in Table VII and Table VIII. We can observe that we can divide the previous works into two big groups, those based on hand-engineered representations and deep learning methods. In general, deep learning methods outperform methods that use handcrafted features. However, their black-box nature could result in decreased interpretability, missing cues that show where and when manifestations of depression are seen, something that could make them more useful as tools for medical practitioners.

Tables VII and VIII show, respectively, the performance of several of these methods on AVEC2013 and AVEC2014, both for (data) monomodal and multimodal approaches. The results of these methods seem to improve when using a multimodal approach with different feature modalities [38] where geometric and texture features are combined. Our proposed method builds on similar ideas, but combines novel physiological features with typical dynamic texture features to exploit mostly the complementary visual and physiological temporal information provided by each subject. The learning based methods mostly rely on exploiting also the temporal information using different different deep learning architectures that search for temporal cues in the stream of frames, potentially exploiting spatio-temporal relationships in the videos that could be indicative of depression.

For AVEC2013, the proposed modality in this study outperforms the hand-engineering ”traditional” methods, even as a (data) monomodal approach, resulting on a 7.54 MAE. In addition, it has similar performance than one of the first learning-based method proposed to compute the depression level based in two DCNNs [41]. To show that our proposed modality and method extracts complementary information with other approaches based on visual information, we combined our results with other types of features. When our modality is fused with other textural or deep modalities, our results show results comparable (e.g.) to the state-of-the-art methods evaluated in AVEC2013, demonstrating the complementary of the information of both modalities.

TABLE VIII
COMPARISON OF METHODS FOR PREDICTING THE LEVEL OF DEPRESSION ON THE AVEC2014 DATASET.

| Methods | MAE | RMSE |
|---|---|---|
| AVEC 2014 Video Baseline [24] | 8.86 | 10.86 |
| MHH + PLS (Jan *et al.* [44]) | 8.44 | 10.50 |
| LGBP-TOP + LPQ (Kaya *et al.* [45]) | 8.20 | 10.27 |
| Two DCNN (Zhu *et al.* [41]) | 7.47 | 9.55 |
| C3D (Jazaery *et al.* [42]) | 7.22 | 9.20 |
| VGG + FDHH (Jan *et al.* [46]) | 6.68 | 8.04 |
| Four DCNN (Zhou *et al.* [43]) | 6.21 | 8.39 |
| ResNet-50 (Melo *et al.* [34]) | 6.15 | 8.23 |
| 3DCNN + SVR (Niu *et al.* [35]) | **6.14** | 7.98 |
| Two-stream 2DCNN (Melo *et al.* [22]) | 6.20 | **7.94** |
| Ours (HRV) | 7.44 | 9.55 |
| Ours (HRV + LGBP-TOP) | 6.81 | 8.63 |
| Ours (HRV + LGBP-TOP + Deep) | 6.57 | 8.49 |

For AVEC2014, our method, using exclusively the HRV features as the data modality, also outperforms traditional methods using handcrafted features from the RGB videos, and is very close to some deep learning-based methods such as Zhu et al. [41]. When we combine the features derived from the rPPG signal with deep or visual texture-based features, we achieve results comparable to the state-of-the-art methods in the detection of depression. The improvement of modality fusion at the score level is worse than when testing in AVEC2013, probably due to a smaller amount of data.

## IV. Conclusion

This paper introduced the extraction of remote biosignals from RGB videos to be used in automatic screening of depression levels from facial videos, a novel visual data modality explored here for the first time. In this context, we have proposed a novel scheme that directly extracts physiological signals in an unsupervised manner, just based on visual information, removing the need for any contact-based device or reference signal. We have directly used these signals to compute physiological features such as blood volume pulse features or heart rate variability parameters, training different machine learning regression models. We evaluated our approach using the AVEC2013 and 2014 benchmark databases. Our results show that our method provides information that can help in the assessment of depression, proving that it can be combined with other visual data modalities to improve the performance further. In our analysis, we have shown graphical examples that suggest that the inference of the models trained with this type of feature modality is slightly more stable than those of other models, such as those that exploit textural or deep features. Extensive experiments indicated the usefulness of such modality, when compared to different methods present in the literature.

## V. Acknowledgements

This research has been supported by the Academy of Finland 6G Flagship program under Grant 346208 and PROFI5 HiDyn under Grant 326291. The authors wish to acknowledge CSC, IT Center for Scientific, Finland, for computational resources.

## References

[1] J. Verhoeven, D. Révész, J. Lin, O. Wolkowitz, and B. Penninx, "Major depressive disorder and accelerated cellular aging: Results from a large psychiatric cohort study," *Molecular psychiatry*, vol. 19, 11 2013.

[2] A. Pampouchidou, P. Simos, K. Marias, F. Meriaudeau, F. Yang, M. Pediaditis, and M. Tsiknakis, "Automatic assessment of depression based on visual cues: A systematic review," *IEEE Trans. on Affective Computing*, pp. 1–27, 2017.

[3] R. Hartmann, F. M. Schmidt, C. Sander, and U. Hegerl, "Heart rate variability as indicator of clinical state in depression," *Frontiers in Psychiatry*, vol. 9, 2019. [Online]. Available: https://www.frontiersin.org/article/10.3389/fpsyt.2018.00735

[4] S. Byun, A. Y. Kim, E. H. Jang, S. Kim, K. W. Choi, H. Y. Yu, and H. J. Jeon, "Detection of major depressive disorder from linear and nonlinear heart rate variability features during mental task protocol," *Computers in Biology and Medicine*, vol. 112, p. 103381, 2019. [Online]. Available: https://www.sciencedirect.com/science/article/pii/S0010482519302586

[5] M. Kobayashi, G. Sun, T. Shinba, T. Matsui, and T. Kirimoto, "Development of a mental disorder screening system using support vector machine for classification of heart rate variability measured from single-lead electrocardiography," in *2019 IEEE Sensors Applications Symposium (SAS)*, March 2019, pp. 1–6.

[6] M. Sarchiapone, c. m. Gramaglia, M. Iosue, V. Carli, L. Mandelli, A. Serretti, D. Marangon, and P. Zeppegno, "The association between electrodermal activity (eda), depression and suicidal behaviour: A systematic review and narrative synthesis," *BMC Psychiatry*, vol. 18, 01 2018.

[7] C. Álvarez Casado and M. Bordallo López, "Face2ppg: An unsupervised pipeline for blood volume pulse extraction from faces," 2022. [Online]. Available: https://arxiv.org/abs/2202.04101

[8] T. Tamura, "Current progress of photoplethysmography and spo2 for health monitoring," *Biomedical Engineering Letters*, vol. 9, 02 2019.

[9] W. Liu, D. Anguelov, D. Erhan, C. Szegedy, S. E. Reed, C. Fu, and A. C. Berg, "SSD: single shot multibox detector," *CoRR*, vol. abs/1512.02325, 2015. [Online]. Available: http://arxiv.org/abs/1512.02325

[10] M. Kowalski, J. Naruniec, and T. Trzcinski, "Deep alignment network: A convolutional neural network for robust face alignment," *CoRR*, vol. abs/1706.01789, 2017. [Online]. Available: http://arxiv.org/abs/1706.01789

[11] C. Álvarez Casado and M. Bordallo López, "Real-time face alignment: evaluation methods, training strategies and implementation optimization," *Journal of Real-Time Image Processing*, pp. 1–29, 2021.

[12] G. de Haan and V. Jeanne, "Robust pulse rate from chrominance-based rppg," *IEEE Transactions on Biomedical Engineering*, vol. 60, no. 10, pp. 2878–2886, Oct 2013.

[13] C. Álvarez Casado, P. Paananen, P. Siirtola, S. Pirttikangas, and M. Bordallo López, *Meditation Detection Using Sensors from Wearable Devices*. New York, NY, USA: Association for Computing Machinery, 2021, p. 112–116. [Online]. Available: https://doi.org/10.1145/3460418.3479318

[14] C. R. Harris, K. J. Millman, S. J. van der Walt, R. Gommers, P. Virtanen, D. Cournapeau, E. Wieser, J. Taylor, S. Berg, N. J. Smith, R. Kern, M. Picus, S. Hoyer, M. H. van Kerkwijk, M. Brett, A. Haldane, J. F. del Río, M. Wiebe, P. Peterson, P. Gérard-Marchant, K. Sheppard, T. Reddy, W. Weckesser, H. Abbasi, C. Gohlke, and T. E. Oliphant, "Array programming with NumPy," *Nature*, vol. 585, no. 7825, pp. 357–362, Sep. 2020. [Online]. Available: https://doi.org/10.1038/s41586-020-2649-2

[15] F. Shaffer and J. P. Ginsberg, "An overview of heart rate variability metrics and norms," *Frontiers in Public Health*, vol. 5, 2017. [Online]. Available: https://www.frontiersin.org/article/10.3389/fpubh.2017.00258

[16] R. Vallat and M. Walker, "An open-source, high-performance tool for automated sleep staging," *eLife*, vol. 10, 10 2021.

[17] D. Makowski, T. Pham, Z. J. Lau, J. C. Brammer, F. Lespinasse, H. Pham, C. Schölzel, and S. H. A. Chen, "NeuroKit2: A python toolbox for neurophysiological signal processing," *Behavior Research Methods*, vol. 53, no. 4, pp. 1689–1696, feb 2021. [Online]. Available: https://doi.org/10.3758%2Fs13428-020-01516-y

[18] P. van Gent, H. Farah, N. van Nes, and B. van Arem, "Heartpy: A novel heart rate algorithm for the analysis of noisy signals," *Transportation Research Part F: Traffic Psychology and Behaviour*, vol. 66, pp. 368–378, 2019. [Online]. Available: https://www.sciencedirect.com/science/article/pii/S1369847818306740

[19] M. Valstar, B. Schuller, K. Smith, T. Almaev, F. Eyben, J. Krajewski, R. Cowie, and M. Pantic, "Avec 2014 - 3d dimensional affect and depression recognition challenge," *AVEC 2014 - Proceedings of the 4th International Workshop on Audio/Visual Emotion Challenge, Workshop of MM 2014*, pp. 3–10, 11 2014.

[20] T. R. Almaev and M. F. Valstar, "Local gabor binary patterns from three orthogonal planes for automatic facial expression recognition," in *2013 Humaine Association Conference on Affective Computing and Intelligent Interaction*, Sep. 2013, pp. 356–361.

[21] A. Anjos, M. Günther, T. de Freitas Pereira, P. Korshunov, A. Mohammadi, and S. Marcel, "Continuously reproducing toolchains in pattern recognition and machine learning experiments," in *International Conference on Machine Learning (ICML)*, Aug. 2017. [Online]. Available: http://publications.idiap.ch/downloads/papers/2017/Anjos_ICML2017-2_2017.pdf

[22] W. Carneiro de Melo, E. Granger, and M. Bordallo Lopez, "Encoding temporal information for automatic depression recognition from facial analysis," in *ICASSP 2020 - 2020 IEEE International Conference on Acoustics, Speech and Signal Processing (ICASSP)*, May 2020, pp. 1080–1084.

[23] M. Valstar, B. Schuller, K. Smith, F. Eyben, B. Jiang, S. Bilakhia, S. Schnieder, R. Cowie, and M. Pantic, "Avec 2013 - the continuous audio/visual emotion and depression recognition challenge," 10 2013, pp. 3–10.

[24] M. V. *et al.*, "Avec 2014: 3d dimensional affect and depression recognition challenge," in *AVEC 2014*, 2014, pp. 3–10.

[25] A. Jan, H. Meng, Y. F. Abdul Gaus, and F. Zhang, “Artificial intelligent system for automatic depression level analysis through visual and vocal expressions,” *IEEE Transactions on Cognitive and Developmental Systems*, vol. PP, pp. 1–1, 07 2017.

[26] M. Niu, J. Tao, B. Liu, J. Huang, and Z. Lian, “Multimodal spatiotemporal representation for automatic depression level detection,” *IEEE Transactions on Affective Computing*, vol. PP, pp. 1–1, 10 2020.

[27] Y. Dong and X. Yang, “A hierarchical depression detection model based on vocal and emotional cues,” *Neurocomputing*, vol. 441, pp. 279–290, 2021. [Online]. Available: https://www.sciencedirect.com/science/article/pii/S0925231221002654

[28] C. Cai, M. Niu, B. Liu, J. Tao, and X. Liu, “Tdca-net: Time-domain channel attention network for depression detection,” 08 2021, pp. 2511–2515.

[29] M. S’adan, A. Pampouchidou, and F. Meriaudeau, “Deep learning techniques for depression assessment,” 08 2018.

[30] S. Song, S. Jaiswal, L. Shen, and M. Valstar, “Spectral representation of behaviour primitives for depression analysis,” *IEEE Transactions on Affective Computing*, pp. 1–1, 2020.

[31] S. Rathi, B. Kaur, and R. Agrawal, “Selection of relevant visual feature sets for enhanced depression detection using incremental linear discriminant analysis,” *Multimedia Tools and Applications*, 03 2022.

[32] L. Wen, X. Li, G. Guo, and Y. Zhu, “Automated depression diagnosis based on facial dynamic analysis and sparse coding,” *IEEE Transactions on Information Forensics and Security*, vol. 10, no. 7, pp. 1432–1441, July 2015.

[33] Y. Zhu, Y. Shang, Z. Shao, and G. Guo, “Automated depression diagnosis based on deep networks to encode facial appearance and dynamics,” *IEEE Transactions on Affective Computing*, vol. PP, pp. 1–1, 01 2017.

[34] W. C. de Melo, E. Granger, and A. Hadid, “Depression detection based on deep distribution learning,” in *2019 IEEE International Conference on Image Processing (ICIP)*, Sep. 2019, pp. 4544–4548.

[35] M. Niu, J. Tao, and B. Liu, “Multi-scale and multi-region facial discriminative representation for automatic depression level prediction,” in *ICASSP 2021 - 2021 IEEE International Conference on Acoustics, Speech and Signal Processing (ICASSP)*, June 2021, pp. 1325–1329.

[36] D. Ahmad, R. Goecke, and J. Ireland, *CNN Depression Severity Level Estimation from Upper Body vs. Face-Only Images*, 02 2021, pp. 744–758.

[37] H. M. *et al.*, “Depression recognition based on dynamic facial and vocal expression features using partial least square regression,” in *AVEC 2013*, 2013, pp. 21–30.

[38] H. Kaya and A. A. Salah, “Eyes whisper depression: A cca based multimodal approach,” in *Proceedings of the 22nd ACM International Conference on Multimedia*, ser. MM ’14. New York, NY, USA: Association for Computing Machinery, 2014, p. 961–964. [Online]. Available: https://doi.org/10.1145/2647868.2654978

[39] M. K. *et al.*, “Fusion of audio-visual features using hierarchical classifier systems for the recognition of affective states and the state of depression,” in *ICPRAM 2014*, 2014, pp. 671–678.

[40] L. Wen, X. Li, G. Guo, and Y. Zhu, “Automated depression diagnosis based on facial dynamic analysis and sparse coding,” *IEEE Trans. on Information Forensics and Security*, vol. 10, pp. 1432–1441, 2015.

[41] Y. Zhu, Y. Shang, Z. Shao, and G. Guo, “Automated depression diagnosis based on deep networks to encode facial appearance and dynamics,” *IEEE Trans. on Affective Computing*, vol. 9, no. 4, pp. 578–584, 2018.

[42] M. Jazaery and G. Guo, “Video-based depression level analysis by encoding deep spatiotemporal features,” *IEEE Trans. on Affective Computing*, pp. 1–8, 2018.

[43] X. Zhou, K. Jin, Y. Shang, and G. Guo, “Visually interpretable representation learning for depression recognition from facial images,” *IEEE Trans. on Affective Computing*, pp. 1–12, 2018.

[44] A. Jan, H. Meng, Y. Gaus, F. Zhang, and S. Turabzadeh, “Automatic depression scale prediction using facial expression dynamics and regression,” in *AVEC 2014*, 2014, pp. 73–80.

[45] H. Kaya, F. Çilli, and A. Salah, “Ensemble cca for continuous emotion prediction,” in *AVEC 2014*, 2014, pp. 19–26.

[46] A. Jan, H. Meng, Y. Gaus, and F. Zhang, “Artificial intelligent system for automatic depression level analysis through visual and vocal expressions,” *IEEE Trans. on Cognitive and Developmental Systems*, vol. 10, pp. 668–680, 2018.